\documentclass{article}

\usepackage{lineno,hyperref}
\modulolinenumbers[5]
\usepackage{arxiv}
\usepackage{amsmath,amsthm}
\usepackage[utf8]{inputenc} 
\usepackage[T1]{fontenc}    
\usepackage{hyperref}       
\usepackage{url}            
\usepackage{booktabs}       
\usepackage{amsfonts}       
\usepackage{nicefrac}       
\usepackage{microtype}      
\usepackage{lipsum}		
\usepackage{graphicx}
\usepackage{natbib}
\usepackage{doi}
\usepackage{algorithmic}
\usepackage{textcomp}
\usepackage[ruled,vlined]{algorithm2e}
\usepackage{caption}
\usepackage{subcaption}
\usepackage{listings}
\usepackage{multirow}
\usepackage{multicol}
\usepackage{pgfplotstable}
\usepackage{pgfplots}
\title{Training a Bidirectional GAN-based One-Class Classifier for Network Intrusion Detection}

\author{ {Wen Xu} \\
	Cybersecurity Lab\\
	Massey University\\
	Auckland, NEW ZEALAND \\
	\texttt{w.xu2@massey.ac.nz} \\
	\And
	{Julian  Jang-Jaccard} \\
	Cybersecurity Lab\\
	Massey University\\
	Auckland, NEW ZEALAND \\
	\texttt{j.jang-jaccard@massey.ac.nz} \\
	\And
	{Tong Liu} \\
	Cybersecurity Lab\\
	Massey University\\
	Auckland, NEW ZEALAND \\
	\texttt{T.Liu@massey.ac.nz}
	\And
	{Fariza Sabrina} \\
	School of Engineering and Technology\\
	CQ University\\
	Sydney, AUSTRALIA \\
	\texttt{f.sabrina@cqu.edu.au} \\
}

\hypersetup{
pdftitle={A template for the arxiv style},
pdfsubject={q-bio.NC, q-bio.QM},
pdfauthor={David S.~Hippocampus, Elias D.~Striatum},
pdfkeywords={First keyword, Second keyword, More},
}

\begin{document}

\maketitle

\begin{abstract}
Existing generative adversarial networks (GANs), primarily used for creating fake image samples from natural images, demand a strong dependence (i.e., the training strategy of the generators and the discriminators require to be in sync) for the generators to produce as realistic fake samples that can “fool” the discriminators. We argue that this strong dependency required for the natural image datasets often doesn’t work in imbalanced text-based datasets such as network intrusion traffic samples where the semantic relationship across text features handled by the generator and the discriminator differ. To address this issue, we propose a new training strategy based on Bidirectional GAN (Bi-GAN) where the requirement to be in sync in the training strategy of the generator and discriminator is relaxed. In our proposed method, the training iteration of the generator (and accordingly the encoder) is increased separate from the training of the discriminator until it satisfies the condition associated with the cross-entropy loss. Our empirical results show that this proposed training strategy greatly improves the performance of both the generator and the discriminator even in the presence of imbalanced classes. In addition, our model offers a new construct of a one-class classifier using the trained encoder-discriminator. The one-class classifier detects anomalous network traffic based on binary classification results instead of calculating expensive and complex anomaly scores (or thresholds). Our experimental result illustrates that our proposed method is highly effective to be used in network intrusion detection tasks and outperforms other similar generative methods on the NSL-KDD dataset.
\end{abstract}


\section{Introduction}
\label{sec:introduction}
Network intrusion detection is used to discover any unauthorized attempts to a network by analyzing network traffic coming in and out of the network looking for any signs of malicious activity \cite{jang2014survey}.  This is often regarded as one of the most important network security mechanisms to block or stop cyberattacks.

Traditional machine learning (ML) approaches, such as supervised network intrusion detection, have shown reasonable performance for detecting malicious payloads included in network traffic-based data sets labeled with ground truth \cite{ett.4150}. However, with the mass increase in the size of data, it has become either too expensive or no longer possible to label a huge number of data sets (i.e., big data) \cite{zhu2021joint}. Unsupervised intrusion detection methods have been proposed as it no longer demands the requirement for labeled data. In addition, these unsupervised methods can utilize only the samples from one class (e.g., normal samples) for training to recognize any patterns that deviate from the training observations. However, the detection accuracy of these unsupervised learning methods tends to suffer as soon as an imbalanced class appears (e.g., the number of samples of a class is significantly more or less compared to the number of samples in other classes).

A number of generative models have been proposed including Autoencoders  \cite{kingma2013auto} and generative adversarial networks (GANs) \cite{goodfellow2014generative} with the ability to generate realistic synthetic data sets or to use these generative capabilities to address other issues such as anomaly detection.

Autoencoder (AE) is composed of an encoder and decoder. The encoder can compress high-dimensional input data into low-dimensional latent space. The decoder generates the output that resembles the input by reassembling the (reduced) data representation from the latent space. Typically, a reconstruction loss is computed between the output and the input and used as a mechanism to identify anomalies. The encoder in AE captures the semantic attributes of the input data into the latent space, as a form of vector representation,  to represent the corresponding input sample from the real data space.

In particular, GANs have emerged as a leading yet very powerful technique for generating realistic data sets especially in the image identification and processing involved in the natural images despite arbitrarily complex data distributions that exist in the real data sets. In GANs, two deep neural networks, the generator and the discriminator respectively, are involved in the main GANs’ structure. The generator and the discriminator are trained in turn by playing an adversarial game. That is, the generator produces the output (i.e., somewhat based on the distribution of real data) while the discriminator takes the fake data (i.e., the output of the generator) and real data as input and aims to distinguish them. The goal of the generator is to generate the fake data that resembles as much as the real data to “fool” the discriminator (i.e., it can’t distinguish the fake data from real data). The generator and the discriminator are highly dependent on each other to reach the optima. That is, the number of training iteration both the generator and the discriminator go through are typically in sync. That is because only a better generator can fool the discriminator successfully thus motivates the discriminator to improve its ability to detect fake data. Similarly, a better discriminator can stimulate the generator more to produce fake data that resembles the real data as much as possible. 

This training strategy works well in some tasks, such as data argumentation involved in producing more synthesized images from real images, where the generators and discriminators can be in sync in their training approaches. It often does not work well in the tasks involved in text-based data processing such as network intrusion detection tasks where working with the feature sets whose semantic representation differs across the generator and the discriminator. To address this issue, we propose a new training strategy based on Bidirectional GAN (Bi-GAN).

The contribution of our proposed approach is following:

\begin{itemize}
	\item We relax the requirement for the generator and the discriminator to train in sync. The generator (along with the encoder) in our proposed model goes through more rigorous training iterations in order to produce more reliable synthetic data set that highly resembles the real traffic samples that maintain as much of the semantic relations that exist in the feature sets in the real traffic samples. Our proposed model shows that the generator’s performance is greatly improved by offering the generator (and encoder) to train more than the discriminator, which in turn, also actually improves the discriminator’s performance better.
	
	\item In our promised model, a cross-entropy is used to keep track of the overall balance in terms of the number of relative training iterations required for the generator and the discriminator. In addition, we employ a -log (D) trick to train the generator to obtain sufficient gradient in the early training stage by inverting the label. 
	
	\item We offer new construction of a one-class classifier using the trained encoder-discriminator for detecting anomalous traffic from normal traffic instead of having to calculate either anomaly scores or thresholds which are computationally expensive and complex.
	
	\item Our experimental result shows that our proposed method is highly effective in using a GAN-based model for network anomaly detection tasks by achieving more than 92\% F1-score.
\end{itemize}

The rest of this paper is organized as follows. Section \ref{sec:rw} summarizes the review of the literature relevant to our study. Background knowledge in the generic GAN and BiGAN require to understand our study is present in Section \ref{sec:bg}. Section \ref{sec:our_model} describes the details of our proposed model. Section \ref{sec:data} describes the data and the data pre-processing methodologies we used while Section \ref{sec:experiments} demonstrates our experimental results with analysis. Section \ref{sec:conclusion} provides the concluding remarks along with the future work planned.

\section{Related Work}
\label{sec:rw}
In this section, we mainly review the usage of two unsupervised models, namely autoencoder and GAN, in network intrusion detection.


The autoencoder-based approaches typically use reconstruction methods where computing a reconstruction error is used to detect whether something is normal or anomalous \cite{deep2,an2015variational,10.1007/978-3-030-58555-6_20, 9552882}. In this approach, an autoencoder is trained only with the normal samples to learn the distribution in their latent representation and use the distribution to reconstruct the input. This approach enables an autoencoder to detect anomalies through the reconstruction error between the original samples and their reconstructions where the non-normal samples typically generate high reconstruction loss compared to the normal samples. In the training phase of autoencoder, no anomalous data or label is required. This makes the autoencoder-based approaches suitable for the domain where there is a sufficient number of normal data obtainable while collecting a sufficient number of anomalous data is difficult and expensive (i.e., network intrusion detection). The work suggested by \cite{9189883} used autoencoder to gain 88.98\% accuracy in network intrusion detection based on the NSL-KDD dataset. In\cite{7987197}, the authors used Denoising Auto-encoders and achieves high performance on the same dataset. Variational Autoencoder (VAE) assumes that the latent representations of the original data obey a probability distribution (e.g, Gaussian distribution) \cite{doersch2016tutorial}. Based on the assumption, An and Cho (2015) \cite{an2015variational} proposed a VAE based anomaly detection by calculating reconstruction probability instead of reconstruction loss to identify anomalies. 

The success of GANs in various domains has drawn attention to applying the GAN and its extensions in detecting anomalous data samples. Although GAN was first published by Goodfellow {et al.} \cite{goodfellow2014generative} in 2014, GANs had not been used for intrusion detection until 2017 when Schlegl \textit{et al.} proposed their AnoGAN ~\cite{schlegl2017unsupervised}. The difficulty behind the GAN-based approach is with the way of obtaining what to consider anomalous (i.e., anomaly criteria). One approach to using the anomaly criteria is to use an anomaly score. Based on this assumption, AnoGAN proposed two losses to form the anomaly score. Since the generator is trained to synthesize data from real data, the first loss \textit{residual loss} is offered to capture the difference between normal data and its synthetic data. The other loss is offered \textit{discriminator loss} to measure the divergence between the output of the discriminator fed with real data and synthetic data. In the proposal, the \textit{discriminator loss} is computed by comparing the result of the discriminator's intermediate layer rather than the output layer. 

The generic GAN model does not have the ability to map from data to latent space thus it is expensive to find the best features in latent space which can generate the most suitable data in computing the \mbox{\textit{residual loss}}. To address this, extra backpropagation is used to find the best features in the generic approach. Schlegl \textit{et al.} further proposed f-AnoGAN \cite{SCHLEGL201930} to improve the computational efficiency by adding an encoder before the generator to map from data to the latent space. The \textit{residual loss} is then replaced with the reconstruction loss of the autoencoder composed by the encoder and the generator. Another GAN-based intrusion detection by the name of GANomaly \cite{10.1007/978-3-030-20893-6_39} adds two encoders in the classic GAN model. In their approach, one encoder is used before the generator to learn the latent space of the original data while the other encoder is used after the generator to learn the latent space of the reconstructed data. With that approach, they offer three losses to measure: \textit{reconstruction loss}, \textit{discriminator loss}, and \textit{encoder loss} which measures the difference between two latent spaces.

BiGAN is a widely used GAN variant that has the ability of mapping data to latent space. BiGAN also comprises three components as f-AnoGAN but has a different structure. In BiGAN, the encoder is independent of the generator as another input source for the discriminator. In this way, the BiGAN can learn to map the latent space to data (by the generator) and vice versa (by the encoder) simultaneously. Efficient GAN \cite{zenati2018efficient} utilized BiGAN's ability of inverse learning and introduced two similar losses as f-AnoGAN to form the anomaly score function. Since the encoder and the generator can compose an auto-encoder structure, \cite{KAPLAN2020185} suggests adding an Autoencoder style training step to the original BiGAN architecture to stabilize the model. Instead of using “feature-matching loss” \cite{zenati2018efficient} as the \textit{discriminator loss} which is computationally expensive, they experimented on the cross-entropy loss between the result of discriminator on the real sample (from the encoder) and 1 (target label) and reported a competitive performance on the test done on the KDD99 data set. 

Using an anomaly score function to detect an anomalous sample is dominant in GAN-based detection tasks. However, we argue that one of the drawbacks of these existing anomaly score-based approaches is that they comprise at least two or even more loss metrics and are generally complex and computation-intensive. A more straightforward resolution is to use a discriminator as a one-class classifier to detect anomalies instead of computing anomaly scores. Mohammadi \textit{et al.}\cite{8990759} constructed an f-AnoGAN like architecture but using discriminator as a one-class classifier for anomaly detection. Their approach was applied to the NSL-KDD dataset and gained performance at the test phase with 91.39\% accuracy.

\section{Background}\label{sec:bg}
\subsection{Generic GAN}
\begin{figure} [h]
	\includegraphics[width=0.45\textwidth]{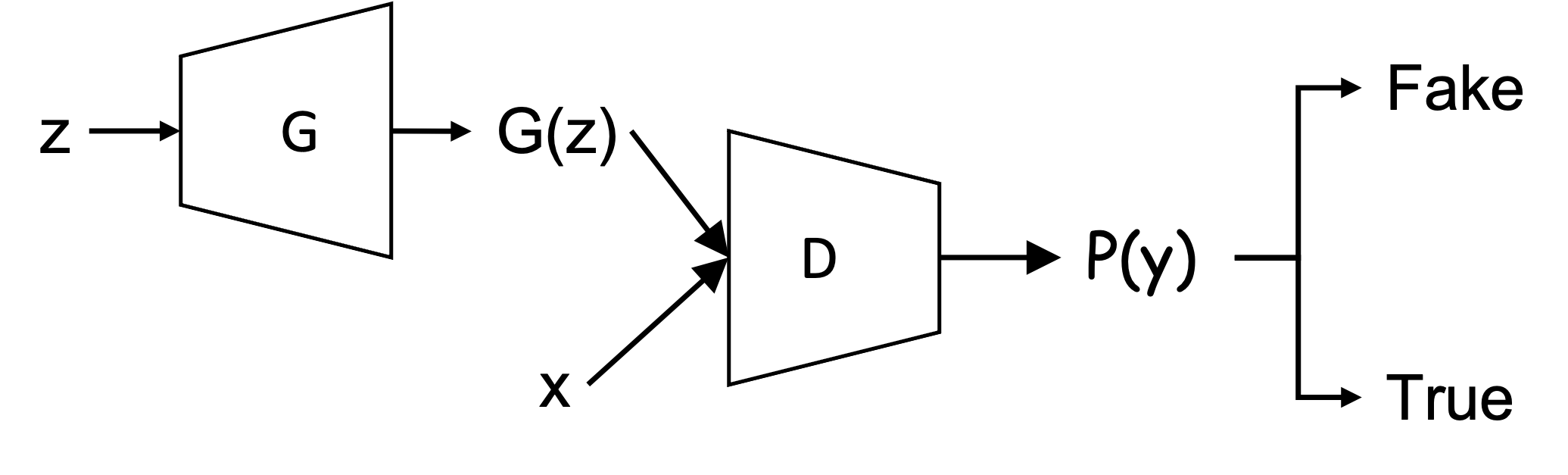}
	\caption{Structure of GAN. The \textit{generator} G map the input \textit{z} (i.e., random noise) in latent space to produce a high dimensional \textit{G(z)} (i.e., fake samples). The \textit{Discriminator} D is expected to separate \textit{x} (i.e., real samples) from \textit{G(z)}.}  	
	\vspace{-8pt}
	\label{fig:GAN}
\end{figure}

In a generic GAN approach, two neural networks, namely the generator and the discriminator respectively, contest with each other in a game approach - generally in the form of a zero-sum game where one agent’s gain is another agent’s loss.
The generative network (i.e., the generator) generates new data samples from a low dimensional distribution while the discriminative network (i.e., the discriminator) evaluates them. In another word, the generator learns to map from random noise to a data distribution of the real data while the discriminator distinguishes the new datasets generated by the generator (i.e., regarded as fake data) from the true data distribution. Fig \ref{fig:GAN} illustrates the architecture of the GAN.

Algorithm \ref{GAN algorithm}  demonstrates the procedure involved in the training phase of a generic GAN model. The generator takes a batch of vectors z (e.g., randomly drawn from a Gaussian distribution) and maps to G(z) which has the same dimensions as the real samples x. The discriminator receives two sources of input (i.e., fake samples and real samples) and tries to distinguish them. The loss between the observation and the prediction at the discriminator is calculated and subsequently used to update both the generator and the discriminator until the training is complete.

\begin{algorithm} [h]
	\SetAlgoLined
	\caption{Training in Generic GAN           	\label{GAN algorithm}}
	\For {number of training iterations} {
		\For {$k$ steps} {
			Sample z (${z^1},{z^2}....{z^n}) \sim p(Z)$\;
			Sample x (${x^1},{x^2}....{x^n}) \in p(X)$\;
			$\hat{x} = G(z)$\;
			Update $D(x, \hat{x})$ by maximizing equation \ref{gan_loss}\;
		}
		Sample z (${z^1},{z^2}....{z^n}) \sim p(Z)$\;
		Update $G(z)$ by minimizing equation \ref{gan_loss} (without updating D).
	}
\end{algorithm}

It must note that there is no independent loss function for the generator in the standard GAN as it is updated indirectly with the objective function linked to the discriminator. Equation \ref{gan_loss} depicts the objective of V(G,D) for measuring the residual and optimizing both the generator and the discriminator.
\begin{multline} \label{gan_loss}
	\underset{G}{min}\ \underset{D}{max}\ V(G, D) =\\
	\mathbb{E}_{x\sim pX}[\log D(x)] + \mathbb{E}_{z\sim pZ}[\log(1-D(G(z)))]
\end{multline}

\subsection{Bidirectional GAN}
\begin{figure} [h]
	\includegraphics[width=0.45\textwidth]{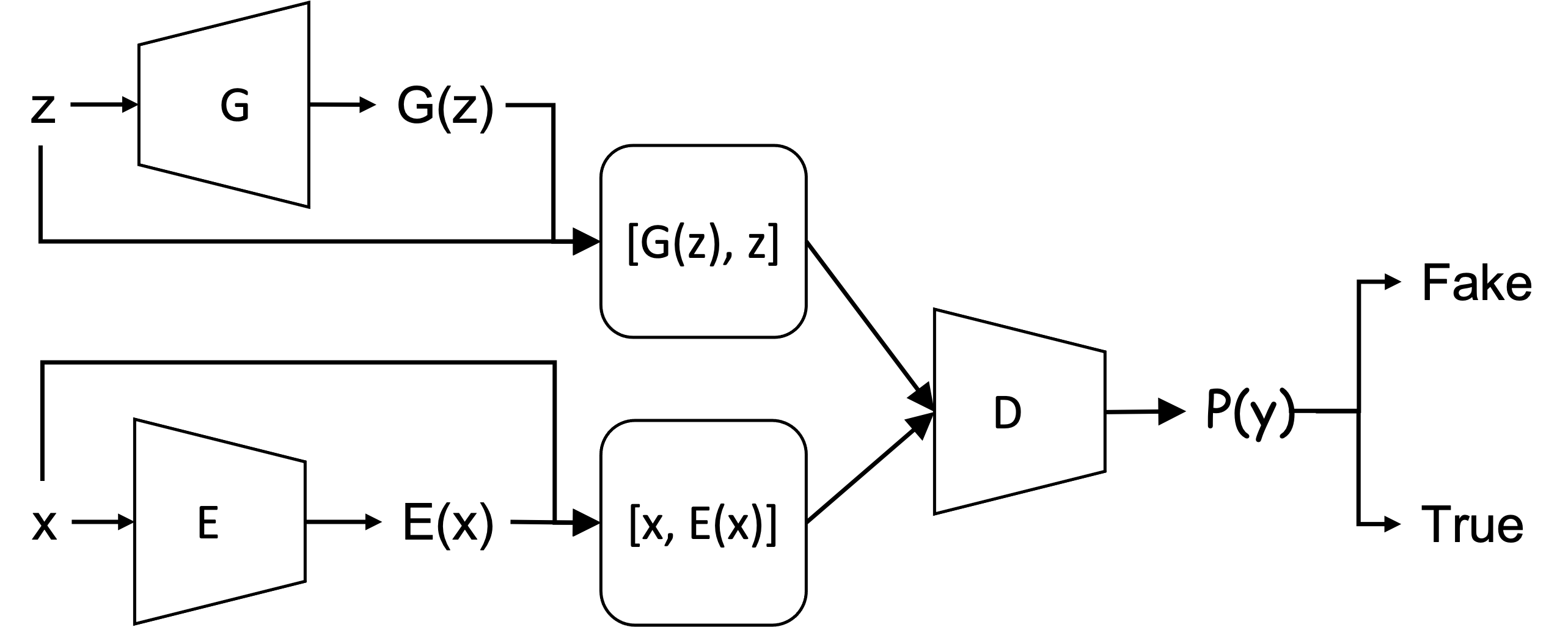}
	\caption{Structure of BiGAN. Note that (z and E(x)) and (G(z) and x) have the same dimensions. The concatenated pairs [G(z), z] and [x, E(x)] are the two input sources of the \textit{discriminator} D. The \textit{Generator} G and the \textit{encoder} E are optimized with the loss generated by the \textit{discriminator} D.}      
	\vspace{-8pt}
	\label{fig:BiGAN}
\end{figure}

Bidirectional GAN or BiGAN is a variant of GAN by adding an encoder to the original GAN model. With the added encoder, the BiGAN is capable to learn the inverse mapping from the real data to the latent space \cite{dumoulin2016adversarially, donahue2017adversarial} to better support the generator producing more semantically rich synthetic datasets. The encoder here plays an importable role for the BiGAN model by providing the learning the latent representation from the real data \cite{dumoulin2016adversarially}. 
Fig \ref{fig:BiGAN} illustrates the architecture of BiGAN.

The algorithm \ref{BiGAN algorithm} depicts the training involved in the BiGAN approach. Like the standard GAN, the training is comprised of two steps. The first step involves training the discriminator (\textit{D}) to maximize the objective function described in Equation \ref{BiGAN_loss} without updating the generator (\textit{G}) or encoder (\textit{E}). The second step involves training both the generator and encoder to minimize the same objective function linked to the discriminator (\textit{D}). 
\begin{multline}
	\min\limits_{G,E}\max\limits_{D}L(D,E,G) = \mathbb{E}_{x\sim pX} [\mathbb{E}_{z\sim pE(\cdot|x)}[\log D(x,z)]] \\ + \mathbb{E}_{z\sim pZ} [\mathbb{E}_{x\sim pG(\cdot|z)}[\log(1- D(x,z))]]
	\label{BiGAN_loss}
\end{multline}

Though the generator (and accordingly encoder) and the discriminator are trained separately, there is only one single objective function computed by the discriminator \textit{D}. Both the generator and the encoder are updated indirectly through discriminator. Different from the standard GAN approach, a concatenation operation is added in the discriminator. The concatenation is to link the data \mbox{\textit{x}} (or \mbox{\textit{G(z)}}) with its latent space \mbox{\textit{E(x)}} (or \mbox{\textit{z}}) and these two concatenated data are then inputted into the discriminator. When optimized, the \textit{G} and the \textit{E} are the inverses mapping to each other \cite{dumoulin2016adversarially, donahue2017adversarial} which can be shown as ${x = G(E(x)}$ and \textit{z = E(G(z)}.

\begin{algorithm} [h]
	\SetAlgoLined
	\caption{Training in BiGAN 	\label{BiGAN algorithm}} 
	\For {number of training iterations} {
		\For {$k$ steps} {
			Sample z (${z^1},{z^2}....{z^n}) \sim p(Z)$\;
			Sample x (${x^1},{x^2}....{x^n}) \in p(X)$\; 
			$f(z) = G(z)$\tcc*[r]{ $f(z).shape = x.shape$}
			$\hat{f}(x) = E(x)$\tcc*[r]{ $\hat{f}(x).shape = z.shape$}
			Concatenate ($[f(z), z]$)\;
			Concatenate ($[x, \hat{f}(x)]$)\;
			Update $D([f(z), z])$ and $D([x, \hat{f}(x)]$ by maximizing equation \ref{BiGAN_loss}\;
		}
		Sample z (${z^1},{z^2}....{z^n}) \sim p(Z)$\;
		Sample x (${x^1},{x^2}....{x^n}) \in p(X)$\; 
		$f(z) = G(z)$\;
		$\hat{f}(x) = E(x)$\;
		Concatenate ($[f(z), z]$)\;
		Concatenate ($[x, \hat{f}(x)]$)\;
		Update $G(z)$ and $E(x)$ simultaneously by minimizing equation \ref{BiGAN_loss} (without updating D).
	}
\end{algorithm}

\section{Our Proposed Model} \label{sec:our_model}
Extending from the BiGAN approach, our proposed model offers a new training strategy for the generator and encoder. By further relaxing the dependence with the discriminator, our proposed model allows the generator and encoder to train until they produce a set of new data samples that resembles the real distribution of the original data while maintaining strong semantic relationships that exist in the text-based features of the original network traffic samples. Our proposed model also offers a new construct for the trained encoder-discriminator to use as a one-class binary classifier. 

\subsection{Main Components} 

\subsubsection{Encoder}
The encoder in our proposed model is used for learning feature representation~\cite{donahue2017adversarial} which takes the real samples as inputs and maps them to a low dimensional vector in a latent space. 
In our approach, the encoder is a neural network with 3 Dense layers and has ReLU as the activation function for the hidden layer and the output layer. 
As this typically works as an inverse mapping of the generator, the size of the latent space is typically set at the same dimension size of the input data used for the generator.

\subsubsection{Generator}
The generator in our approach maps a low-dimensional vector (i.e., random input values) to a higher-dimensional vector in the latent space. The generator acts exactly opposite to the encoder. It accepts a \textit{n}-dimensional noise as the input source and the \textit{n} is identical to the dimension size of latent space in the encoder. 
Our generator draws the \textit{n}-dimensional noise from the standard normal distribution. 
The generator has a neural network structure of 3 Dense layers. ReLU is used as the activation function for the hidden layer while the sigmoid function is used as the activation function for the output layer to restrain the distribution of output within the range of [0,1]. The output layer of the generator has the same number of neurons as the input layer of the encoder and the sigmoid function ensures that the generator's output has the same data distribution range as the encoder's input. 

\subsubsection{Discriminator}
The discriminator in our approach is to distinguish whether the input data is derived from the encoder or forged by the generator in the training phase. It comprises a concatenate layer which receives [x, E(x)] (i.e., the paired input from the encoder) and [G(z), z] (i.e., the paired input from the generator), a hidden dense layer with ReLU as the activation function, and an output layer with only one neuron. The sigmoid activation function is used for the output layer to produce the binary classification result. 

\subsection{Training Phase} 
The ultimate goal of the training strategy involved in a GAN approach is for the generator and the discriminator to reach Nash Equilibrium where their chosen training strategies maximize the payoffs (i.e., the generator produces the fake data as resembles as the real data while the discriminator has built up enough knowledge to distinguish the real from fake samples). In the existing GAN approaches dealing with natural images, this often requires both the generator and the discriminator to improve their capabilities at a relatively equivalent speed.

However, this training strategy often doesn’t work in many application scenarios, because often the semantic relationships of the feature sets require to be maintained in the data set produced by the generator, and the data set being operated on in the discriminator differ from each other, which often leaves the training of the generator unstable (i.e., the loss of the generator fluctuates widely) ~\mbox{\cite{arjovsky2017principled}}. 

In many cases, discriminator converges easily at the beginning of training \cite{berthelot2017began}, making the generator never reach its optimum. To address this issue for network intrusion detection tasks, we train the generator (and the encoder accordingly) more iterations than the discriminator. This new training strategy can prevent an optimum discriminator from appearing too early in the training stage 
thus keeping the training to be more balanced between the generator and discriminator.

Our training strategy is depicted in Algorithm~\ref{props_BiGAN} where the training is processed in mini-batch. In the first stage, the discriminator is trained and updated with a batch of real samples (input from the encoder) and a batch of fake samples (input from the generator) in sequence. In the next stage, the discriminator is fixed and the encoder and the generator are trained \textit{k} times. Like it was used in \cite{donahue2017adversarial}, we also adopt the "\textminus log(D) trick"  \cite{goodfellow2014generative}  to train the generator and the encoder more efficiently by flipping the target label between the generator and the encoder. For example, the input from the encoder is labeled as 1 (fake) while the input from the generator is labeled as 0 (real). Now the result of $O(\hat{y}|y)$ is reversed which generates large gradients. This reflects the second part of an iteration in the Algorithm \ref{props_BiGAN} (i.e., the inner for-loop).

\begin{algorithm} [h]
	\SetAlgoLined
	\caption{Training Phase of our proposed method \label{props_BiGAN}}
	\For {number of training iterations} {
		Sample z (${z^1},{z^2}, \dots, {z^n}) \sim p(Z)$\;
		Sample x (${x^1},{x^2}, \dots, {x^n}) \in p(X)$\;
		$f(z) = G(z)$\tcc*[r]{ $f(z).shape = x.shape$}
		$\hat{f}(x) = E(x)$\tcc*[r]{ $\hat{f}(x).shape = z.shape$}
		Concatenate ($[f(z), z]$)\;
		Concatenate ($[x, \hat{f}(x)]$)\;
		Update $D([f(z), z])$ and $D([x, \hat{f}(x)]$ by maximizing equation \ref{BiGAN_loss}\;
		
		\For {$k$ steps} {
			Sample z (${z^1},{z^2}....{z^n}) \sim p(Z)$\;
			Sample x (${x^1},{x^2}....{x^n}) \in p(X)$\;
			$f(z) = G(z)$\;
			$\hat{f}(x) = E(x)$\;
			Concatenate ($[f(z), z]$)\;
			Concatenate ($[x, \hat{f}(x)]$)\;
			Update $G(z)$ and $E(x)$ simultaneously by minimizing equation \ref{BiGAN_loss} (without updating D).
		}
	}
\end{algorithm}

\textbf{Training Loss Function}

In many cases, Kullback-Leibler (KL) divergence or Jensen-Shannon (JS) divergence is used to measure the distance between the joint distribution of $P(x, E(x))$  and $P(G(z), z)$ in many image-based BiGAN variants which produces the optimum, $E = G^{-1}$ and the divergence of 0. This does not work for text-based BiGAN as the divergence of the two joint distributions of $P(x, E(x))$  and $P(G(z), z)$ cannot be computed directly but can only be approximated indirectly using the discriminator. With this understanding, we use the cross-entropy to approximate the divergence. The Equation \ref{Eq:X-entropy} depicts the cross-entropy of two distribution P(x) and Q(x) and where P(x) is the actual target (0 or 1) and the Q(x) is the joint distribution of $P(x, E(x))$ or $P(G(z), z)$. 

\begin{equation}\label{Eq:X-entropy}
	H(P,Q) = \mathbb{E}_{x\sim P}[\log Q(x)] = -\sum_{x=1}^{n}{ }P(x)log(Q(x))
\end{equation}
where P(x) is the actual target (0 or 1) and the Q(x) is the predict. Now we can unify the updating of the encoder, generator, and discriminator to minimize the result of this loss. 

In the first stage of training, the discriminator is updated in iteration: if the input comes from the encoder, the second part of the objective function (\ref{BiGAN_loss}) becomes 0. maximizing the first part of the objective function equals to minimize the cross-entropy values between $P(x=1)$ and $Q(x)$ as seen in the following Equation:
\begin{equation}
	H(1, Q) = -\log D(x, E(x))
\end{equation}

If the input comes from the generator, the first part of the objective function (\ref{BiGAN_loss}) becomes 0. The second part of the objective function is equivalent to minimizing the cross-entropy values between $P(x=0)$ and $Q(z)$,  as seen in the following Equation:
\begin{equation}
	H(0, Q) = - \log D(G(z), z)
\end{equation}

When training the generator and the encoder, the parameters of the discriminator is fixed. In fact, the two modules are trained separately: the encoder is trained in the encoder-discriminator joint structure while the generator is trained in the generator-discriminator joint network.  Since the labels of input are swapped, the target label will set to be 0 when the input is a real sample $x\sim p(x)$. Updating the encoder is to minimize the cross-entropy between $P(x=0)$ and $Q(x)$:
\begin{equation}  \label{eq_encoder}
	H(0, Q) = -\log D(x, E(x))
\end{equation}

On the other hand,  updating the generator is to minimize the cross-entropy between  $P(x=1)$ and $Q(x)$:
\begin{equation} \label{eq_generator}
	H(1, Q) = -\log D(G(z), z)
\end{equation}

\subsection{Testing Phase} \label{test_phase}
After the training is complete, the discriminator has full knowledge of the joint distribution of normal samples. The output probability is close to 0 when the input of the encoder is normal samples, and far from 0 if the input is anomalous samples.

This knowledge at the discriminator is used in the Testing Phase. If the probability value is greater than a given $\kappa$, the discriminator has high confidence to label the test sample anomalous. We follow the convention to set $\kappa = 0.5$ to use it as a marker to decide whether a network traffic record in the testing set is normal or anomalous. In another word, if $O(\hat{y}|y) > 0.5$, the discriminator in our proposed model mark the inputted test data point as an anomaly. This evaluation process is depicted in Algorithm ~\ref{eval_BiGAN}.
\begin{algorithm}
	\SetAlgoLined
	\caption{Testing Phase of our proposed method \label{eval_BiGAN}} 
	\KwIn{
		Test dataset $X =\{x_1,x_2,x_3,\dots, x_n\}$ \\
		Test Label $ Y = \{y_1, y_2,\dots, y_n\}$ \\
		Encoder $E_\phi$; Discriminator $D_\theta$ }
	\KwOut{ $O(\hat{y}|y)$\\}
	\For {$(x, y) \in (X,Y)$} {
		$\eta$ = Concatenate ($[x, E(x)]$)\;
		$O(\hat{y}|y) \leftarrow D_\theta(\eta)$
	}
\end{algorithm}

The intuition behind this one-class classifier is whether a network traffic sample in the test data set is normal or anomalous is following. In the training phase, the encoder only operates on normal data samples with its data distribution represented by $p(x)$ and learns the distribution in the latent representation ($p(E(x)$). Now at the test phase where the encoder receives not only normal data samples but also anomalous data samples ($X'$), the encoder still compresses the anomalous inputs to the same latent distribution ($p(E(x)$). Since the anomalous input samples falls outside the "normal distribution" ($X'\notin p(x)$), their latent representations ($p(E(X')$) are most likely outliers, that is $p(E(X') \notin p(E(x)$. This enables the discriminator to produce a high probability value close to 1, which now the discriminator can mark them as anomalies.

\subsection{Putting it Together}

\begin{figure}[h]
	\centering
	\includegraphics[width=0.48\textwidth]{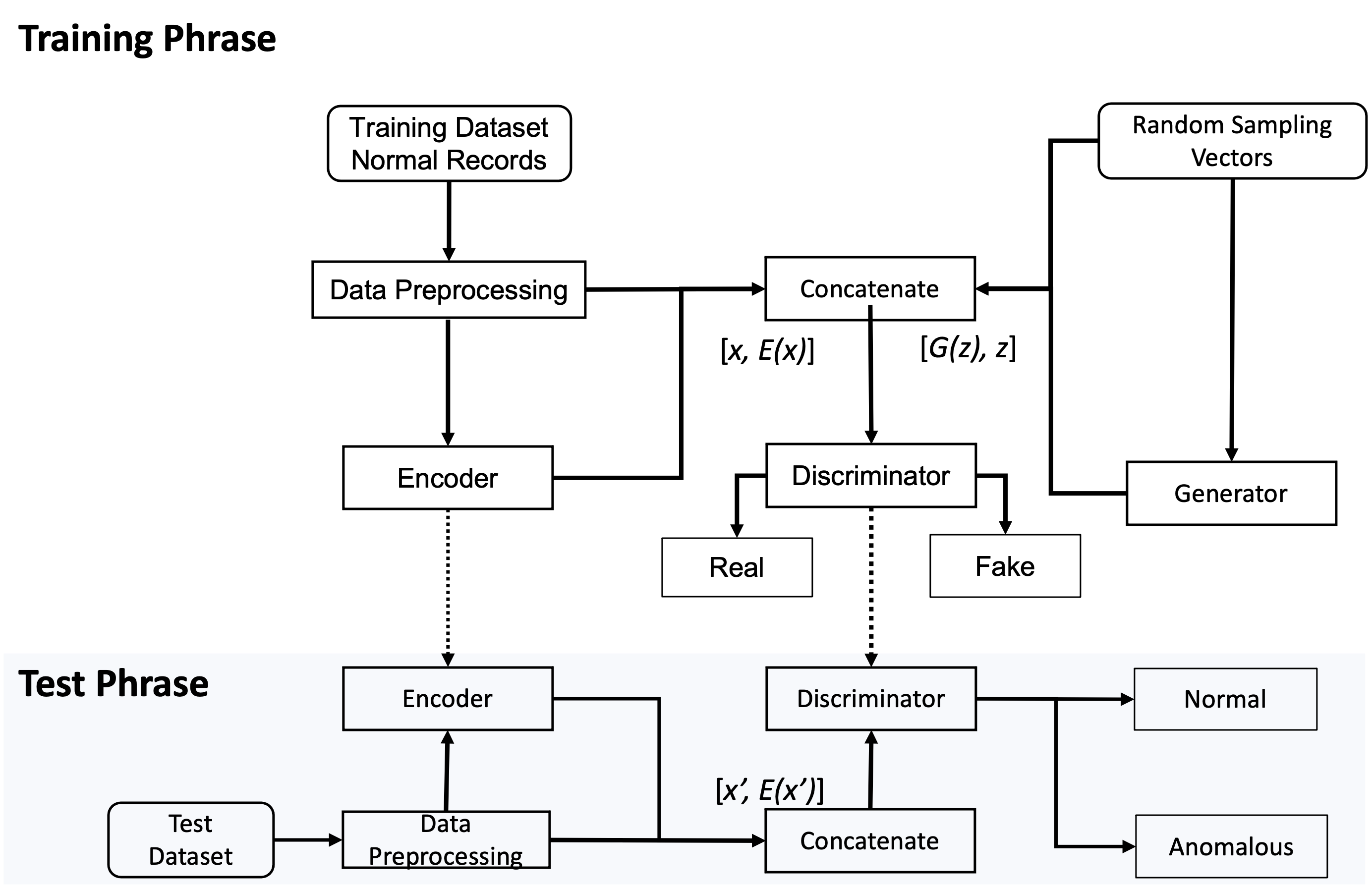}
	\caption{Flowchart of our proposed approach}
	\label{fig:BiGAN_Flowchart}
\end{figure}

Figure \ref{fig:BiGAN_Flowchart} illustrates our proposed approach. 

During the Training Phase, a normal traffic sample $x$ is processed to feed into the encoder as the input. The encoder maps $x$ to latent representations $E(x)$ as the output. Their concatenation $[x, E(x)]$ becomes an input to the discriminator representing real data, labeled with the value 0.

The generator draws a low dimensional vector $z$ and fills it with random values derived from a standard distribution to generate synthetic samples $G(z)$. Their concatenation $[G(z), z]$ becomes another input to the discriminator representing fake data, labeled with the value 1.

The discriminator outputs a probability $O(\hat{y}|y)$ to estimate whether the input is a "real" sample or a "fake" sample. For example, if $O(\hat{y}|y) \approx 0$, the discriminator predicts that the input is the real sample from the encoder, otherwise, the input is the fake sample from the generator. The cross-entropy is used as the unified loss function to update all the three modules in our proposed model while flipping labels provide a strong gradients signal for updating the encoder and the generator. 

During the Testing Phase, the testing samples (x’) containing both normal and abnormal traffic samples are inputted to the encoder which outputs a low dimensional feature representation of these inputs $E(x)$. The paired vector $[x, E(x)]$ becomes the input to the discriminator. The discriminator, by now well trained to see which probability values to associate with the real or fake samples, produces a probability value of each input sample, and predicts whether the input is normal or anomalous since the anomalous input produces considerably dissimilar probabilistic values observed during the training on normal samples.

\section{Data and Pre-processing}\label{sec:data}

\subsection{Datasets}
We use the NSL-KDD dataset for our study. Though the dataset is not a perfect representative of existing real networks, because of the lack of public datasets available for building new models for network-based intrusion detection, it has been widely used as an effective benchmark to compare different intrusion detection methods, along with UNSW-NB15 and CICIDS-2017.

We use two full subsets of the NSL-KDD dataset where the KDDTrain+ contains the dataset that can be used for training a model while KDDTest+ contains the dataset for testing the model. Among the total of 125,973 records in the KDDTrain+ dataset, a total 67,343 of records are considered as normal samples while a total of 58,630 records are categorized as abnormal samples. Similarly, among the total of 22,544 records in the KDDTest+ dataset, the records are grouped into a total of 9,711 normal samples and 12,833 abnormal samples respectively. The details of NSL-KDD is shown in Table \ref{table:nsl_kdd}.

\begin{center}
	\begin{table}[h]
		\setlength{\tabcolsep}{6mm}
		\caption{Records of two NSL-KDD datasets: KDDTrain+ and KDDTest+}
		\label{table:nsl_kdd}
		\begin{tabular}{cccc}
			\hline   
			& \\[-2ex]
			{\textbf{NSL-KDD}} &
			{\textbf{Total}} &
			{\textbf{Normal}} & {\textbf{Others}} \\ 
			\hline
			KDDTrain+ &125,973 &67,343 &58,630\\
			KDDTest+	&22,544	&9,711	&12,833\\
			\hline
		\end{tabular}
	\end{table} 
\end{center}

Each traffic record in the NSL-KDD dataset contains a total of 41 features, including 38 numeric (e.g., "int64" or "float64") and 3 symbolic values (e.g., "object"). Table ~\ref{data_features} shows the details of all 41 features including the name of the feature and data type. 

\begin{center}
	\begin{table} [h]
		\setlength{\tabcolsep}{0.8mm}
		\caption{NSL-KDD dataset features: 38 numeric and 3 symbolic}
		\label{data_features}
		\begin{tabular}{cllcll} 
			\hline
			& \\[-2ex]
			{\textbf{No}} &
			{\textbf{Features}} &
			{\textbf{Type}} & 
			{\textbf{No}} &
			{\textbf{Features}} &
			{\textbf{Type}} \\
			\hline
			0 & duration & int64 & 21	& is\_guest\_login & int64  \\
			1 & protocol\_type & object & 22 & count & int64  \\
			2 & service  &	object 	& 23 &	srv\_count &int64 \\
			3	& flag & object &	24 &	serror\_rate & float64\\
			4 & src\_bytes & int64  &	25	& srv\_serror\_rate  & float64\\
			5 &	dst\_bytes  & int64  &	26 & rerror\_rate & float64\\
			6 &	land     &	int64  	& 27 &	srv\_rerror\_rate  & float64\\
			7 &	wrong\_fragment  &  	int64 & 28 & same\_srv\_rate  &	float64\\
			8 &	urgent  & int64  &	29 & diff\_srv\_rate  & float64\\
			9 &	hot   &	int64  &	30 & srv\_diff\_host\_rate  & float64\\
			10 & num\_failed\_logins  & int64  &	31 & dst\_host\_count & int64\\
			11 & logged\_in  & int64  &	32 & dst\_host\_srv\_count &	int64\\
			12 & num\_compromised & int64 &	33 & dst\_host\_same\_srv\_rate & float64\\
			13 & root\_shell & int64  &	34 & dst\_host\_diff\_srv\_rate  &	float64\\
			14 & su\_attempted & int64 & 35 & dst\_host\_same\_src\_port\_rate &	float64\\
			15 & num\_root & int64 & 36 & dst\_host\_srv\_diff\_host\_rate & float64\\
			16 & num\_file\_creations & int64 & 37 & dst\_host\_serror\_rate & float64\\
			17 & num\_shells & int64 & 38 & dst\_host\_srv\_serror\_rate  & float64\\
			18 & num\_access\_files  & int64 & 39	& dst\_host\_rerror\_rate & float64\\
			19 & num\_outbound\_cmds & int64 & 40 & dst\_host\_srv\_rerror\_rate & float64\\
			20 & is\_host\_login & int64 & & & \\	 	 	 
			\hline
		\end{tabular}
	\end{table} 
\end{center}

\begin{figure}[h]
	\begin{tabular}{cc}
		\includegraphics[width=4cm, height=3cm]{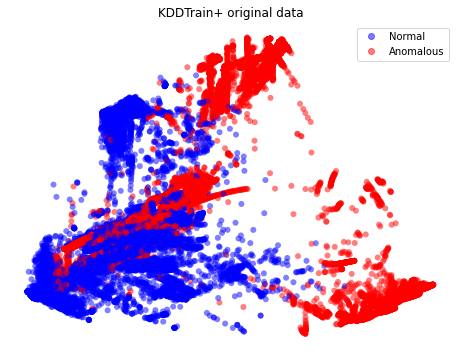} 
		&  
		\includegraphics[width=4cm, height=3cm]{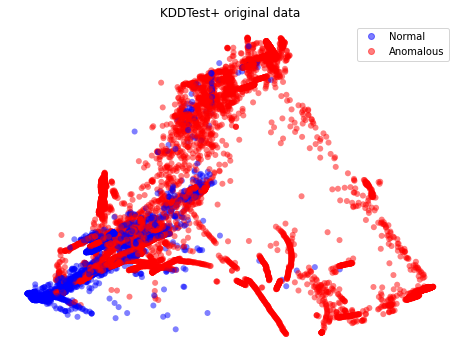}
		\\
		\centering (a)
		& 
		\centering (b) 
	\end{tabular}
	\caption{The PCA visualization of data distribution in KDDTrain+ and KDDTest+ dataset}
	\label{fig:original_pca}
\end{figure}

Fig. \ref{fig:original_pca} illustrates the 2-D visualization of the distribution among the normal and abnormal samples in KDDTrain+ and KDDTest+. As it illustrates, there are two distinct clusters in the KDDTrain+ dataset, one belongs to normal samples and the other belong to abnormal data samples. The feature values are pretty widely spread across each cluster both in the normal and abnormal dataset. In contrast, the clusters around normal and abnormal samples in the KDDTest+ are less distinct as there are many overlapping data points across the normal and abnormal samples. The feature distribution within the normal dataset is within a narrow range while the feature distribution within the abnormal dataset is much wider.

\subsection{Data Pre-processing}
We first encoded 3 symbolic features using one-hot-encoding and converted them into 84 unique features. This resulted in the input source of the encoder being a total of 122 features. After applying one-hot-encoding, outliers are removed following the operation applied in \mbox{\cite{9552882}} – that is, the instances that contain the top 5\% values of any feature (including encoded categorical features) are disposed of. By removing outliers after the one-hot-encoding, all features are now treated equally, regardless of their data types, thus reduce the bias associated with an imbalanced number of particular data types.

As for the input to the generator, the vectors are randomly drawn from a normal distribution of the real dataset to match the size of the latent space dimension (e.g., 10 input vectors according to the latent space with the size of 10). The output size of the generator is decided according to the size of features (e.g., 122 neurons according to the total size of features in the training dataset). As our encoder does the inverse mapping of the generator, the input size corresponds to the size of the total features (e.g., 122) while the output size is the size of the latent space (e.g., 10). The input size of the discriminator is equal to the concatenation of the output of the generator and its input (e.g., 132) while the output size is a single neuron (e.g., as a binary classifier).

\begin{figure}[h]
	\includegraphics[width=0.48\textwidth]{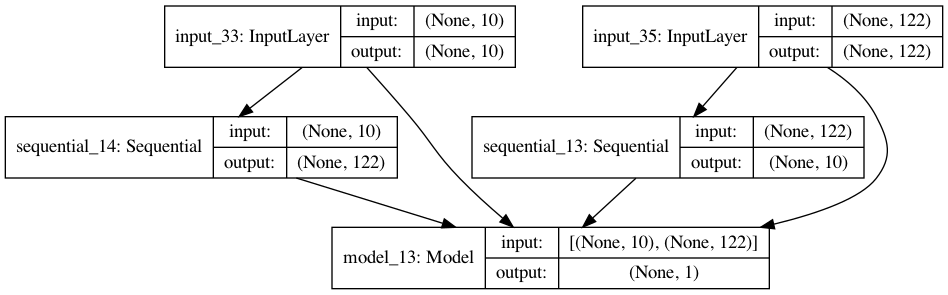}
	\caption{BiGAN in Keras. \textit{Encoder}: input dimensions (122), output dimensions(10); \textit{Generator}: input dimensions (10), output dimensions(122); \textit{Discriminator} concatenates the input and output of \textit{Encoder} or \textit{Generator} to form the input and functions as a binary classifier}       
	\vspace{-8pt}
	\label{fig:model_plot}
\end{figure}

Fig \ref{fig:model_plot} illustrates the relationship between the input and output sizes of the generator, encoder, and discriminator of our proposed model.

\section{Experimental Results}\label{sec:experiments}

\subsection{Setup Environment}
Our experiments were carried out using the following system setup shown in Table~\ref{table:setup}.
\begin{table}[h]
	\centering
	\footnotesize
	\caption{Implementation environment specification}
	\label{table:setup}
	\begin{tabular}{p{2.0cm} | p{3.8cm}}
		\hline
		\textbf{Unit}   & \textbf{Description}\\ \hline
		Processor   & 2.2GH$_z$  Inter Core i7 \\ \hline
		RAM  &  16GB      \\ \hline
		OS  &  MacOS Big Sur   11.4  \\ \hline	
		Packages used  &  tensorflow 2.0.0, sklearn 0.24.1    \\ \hline	
	\end{tabular}
\end{table}

The hyperparameters we used in our study are shown in Table \ref{table:Training parameters}.

\begin{table}[h!]
	\centering
	\caption{Training parameters}
	\begin{tabular}{||l | l | l ||} 
		\hline
		Parameters & Values & Description \\ [0.4ex] 
		\hline\hline
		Batch Size & 64  & The number of training examples in one  \\ 
		&   & forward/backward pass \\ 
		Learning rate & 0.002 & Learning rate is used in the training of \\
		& &  neural networks- range between 0.0\\
		& & and 1.0. \\
		N-iterations & 1000 & Total numbers of iterations in the training \\
		
		&   & process   \\
		Steps & 5 &  The compensated training iterations for \\
		& & Generator and Encoder \\
		\hline
	\end{tabular}
	
	\label{table:Training parameters}
\end{table}

\subsection{Performance Metrics}
We use the classification accuracy, precision, recall, and F1 score as the performance metrics to evaluate the performance effectiveness of our proposed model. We use: True Positive (TP) indicates the number of correctly predicted anomalies, True Negative (TN) indicates the number of correctly predicted normal instances, False Positive (FP) indicates the number of normal instances that are misclassified as anomalies, and False Negative (FN) indicates the number of anomalies that are misclassified as normal.

We use True Positive Rate (also known as Recall) to estimate the ratio of the correctly predicted samples of the class to the overall number of instances of the same class using Equation ~(\ref{eq:TPR}). Typically, the higher TPR $\in [0,1]$ indicates the good performance of the model.

\begin{equation}\label{eq:TPR}
	TPR (Recall) = \frac{TP}{TP + FN}
\end{equation}

We use Precision to measure the quality of the correct predictions which is computed by the ratio of correctly predicted samples to the number of all the predicted samples for that particular class  as seen in Equation~(\ref{eq:PPV}). 

\begin{equation}\label{eq:PPV}
	Precision = \frac{TP}{TP + FP}
\end{equation}

F1-Score computes the trade-off between precision and recall. Mathematically, it is the harmonic mean of precision and recall as shown in Equation ~(\ref{eq:F-measure}).

\begin{equation}\label{eq:F-measure}
	F1 = 2\times\left(\frac{Precision\times Recall}{Precision + Recall}\right)
\end{equation}

Similar to F1-score, we use Accuracy (Acc) to measure the total number of data samples correctly classified in terms of all the predictions made by the model  using Equation~(\ref{eq:ACC}). 

\begin{equation}\label{eq:ACC}
	A{_{CC}} = \frac{TP+TN}{TP + TN + FP + FN}
\end{equation}

We also use the area under the curve (AUC) to compute the area under the receiver operating characteristics (ROC) curve plotting the trade-off between the true positive rate (i.e., typically depicted on the y-axis) and the false positive rate (i.e., on the x-axis across different thresholds) using Equation~(\ref{eq:auc}).

\begin{equation}\label{eq:auc}
	AUC_{ROC}=\int_{0}^{1} \frac{TP}{TP+FN}d\frac{FP}{TN+FP}
\end{equation}

\subsection{Results}
We report the analysis of the observation we have made during our experiments.

\subsubsection{Training}
\begin{figure}[h]
	\centering
	\includegraphics[width=0.40\textwidth]{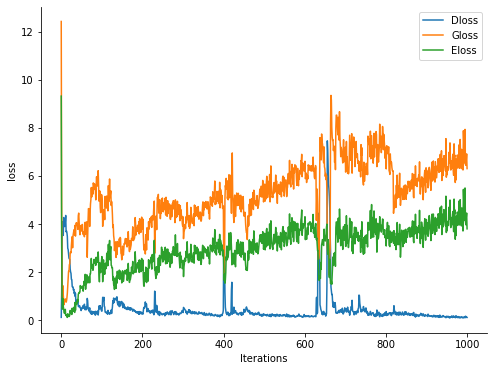}
	\caption{Training Losses vs Iterations, Eloss, Gloss and Dloss represent the training loss trend of encoder, generator and discriminator respectively.}
	\label{fig:losses}
\end{figure}

Fig.~\ref{fig:losses} illustrate the trend of the training losses associated with our main components. In this experiment, the training of the encoder, generator, and discriminator has gone through 1,000 iterations. As expected, the training losses for the encoder (i.e., Eloss depicted by the green line) and generator (i.e., Gloss depicted by the orange line) were not stable in the first 170 iterations but become stead when the training passed 200 iterations. As the encoder is the inverse mapping of the generator, the trend in the training loss patterns is pretty similar between the two. The training loss for discriminator (i.e., Dloss depicted by blue line) becomes stabilized much earlier only after 50 iterations, and remains steady.

\begin{figure}[h]
	\begin{tabular}{cc}
		\includegraphics[width=4cm, height=3cm]{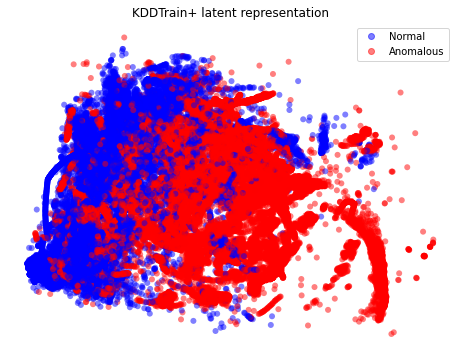} 
		&  
		\includegraphics[width=4cm, height=3cm]{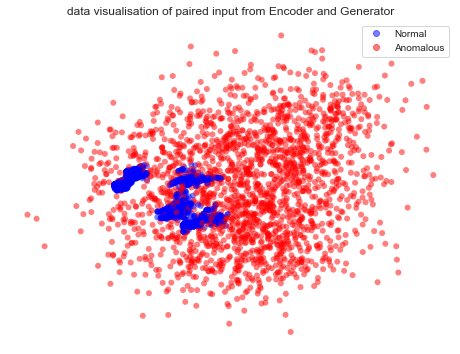}
		\\
		\centering (a)
		& 
		\centering (b)
	\end{tabular}
	\caption{(a) The PCA visualization of data distribution in the latent space of KDDTrain+; (b) The PCA visualization of two paired inputs of discriminator (from encoder and generator)}
	\label{fig:latent_pca}
\end{figure}

Fig.~\ref{fig:latent_pca} illustrates the PCA visualization of two data sets, each representing data distribution in the late space (a) of the original KDDTrain+ data set, (b) of two paired concatenated inputs produced by both the encoder and the generator being feed to the discriminator. As it is shown in graph (a), two clusters around the normal and anomalous traffic samples in KDDTrain+ are noticeably separated from each other though there is a small number of anomalies features mixed within the cluster of the normal samples. As shown in graph (b), there are two distinct clusters around normal and abnormal datasets while the number of the abnormal datasets has significantly increased compared to the original KDDTrain+.  



\subsubsection{Testing}
\begin{figure}[h]
	\centering
	\includegraphics[width=0.4\textwidth]{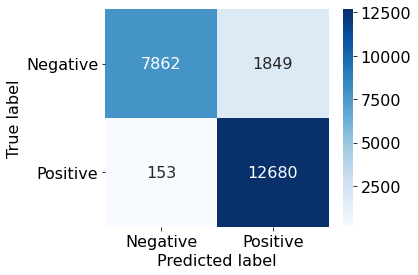}
	\caption{The visualization of latent space representation of KDDTrain+ and KDDTest+}
	\label{fig:confusion_matrix}
\end{figure}

The experimental results of classification accuracy of our proposed model based on the confusion matrix are shown in Fig~\ref{fig:confusion_matrix}. Among the total of 22,544 records used for the testing, 20,542 records were correctly classified according to their label while slightly over 2,000 records (i.e., less than 9\% of the total records) were misclassified as either FP (1,849) or FN (153).

The exact performance of our proposed model is depicted in table~\ref{table:performance}. 
\begin{center}
	\begin{table}[h]
		\setlength{\tabcolsep}{2.0mm}
		\renewcommand{\arraystretch}{1.5}
		\caption{Performance of our apprach on}
		\label{table:performance}
		\begin{tabular}{lrrrr} 
			\hline
			{\textbf{Dataset}} &
			{\textbf{Accuracy}} &
			{\textbf{Precision}}&
			{\textbf{Recall}}&
			{\textbf{F1 score}} \\
			\hline
			KDDTest+ & 91.12\% & 87.27\% & 98.81\% & 92.68\% \\
			\hline
		\end{tabular}
	\end{table} 
\end{center}

\begin{figure}[h]
	\centering
	\includegraphics[width=0.48\textwidth]{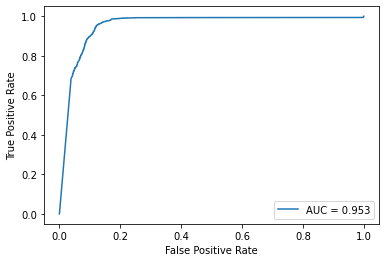}
	\caption{AUC\_ROC curve of our proposed model}
	\label{fig:roc_auc}
\end{figure}

From another angle to measure the performance of our proposed model, Fig~\ref{fig:roc_auc} depicts the AUC\_ROC curve to clearly demonstrate the trade-off between true positive rate and false-positive rate. The curve confirms that our proposed model is highly effective in accurately classifying network intrusions by achieving an AUC\_ROC score at 0.953.

\subsubsection{Benchmarking with other similar models}
We compared the performance of our proposed model with other generative deep neural network models such as Autoencoder and GAN-based approaches test against to similar network intrusion dataset similar to the NSL-KDD dataset (e.g., KDD99). The benchmarking result is shown in Table~\ref{table:performance_benchmark}. Overall, GAN-based approaches show slightly better performance in terms of different performance metrics compared to the Autoencoder-based method. Among different GAN-based methods, our proposed model shows the best performance by achieving more than 92\% F1-score.

\begin{center}
	\begin{table}[h]
		\setlength{\tabcolsep}{1mm}
		\caption{Performance of our approach and other state-of-art approaches}
		\label{table:performance_benchmark}
		\begin{tabular}{lrrrrr} 
			\hline
			{\textbf{Method}} &
			{\textbf{Accuracy}} &
			{\textbf{Precision}}&
			{\textbf{Recall}}&
			{\textbf{F1 score}} &
			{\textbf{Dataset}} \\
			
			\hline
			AE~\cite{IERACITANO202051} & 84.21\% & 87\% & 80.37\% & 81.98\% & NSL-KDD\\
			AE~\cite{9189883} & 88.98\% & 87.92\% & 93.48\% & 90.61\% & NSL-KDD\\
			DAE~\cite{7987197} & 88.65\% & 96.48\& & 83.08\% & 89.28\% & NSL-KDD \\
			AnoGAN~\cite{8594897}   & - & 87.86\% & 82.97\% & 88.65\% & KDD99\\
			BiGAN~\cite{chen2019efficient}  & - & 93.24\%  & 94.73\%  & 93.98\% & KDD99\\
			BiGAN~\cite{KAPLAN2020185} & 89.5\% & 83.6\% & 99.4\% & 90.8\% & KDD99\\
			\textbf{Our approach} & \textbf{91.12\%} & \textbf{87.27\%} & \textbf{98.81\%} & \textbf{92.68\%} & NSL-KDD\\ 
			\hline
		\end{tabular}
	\end{table} 
\end{center}

\section{Conclusion}
\label{sec:conclusion}
In this study, we proposed a new training strategy based on a Bidirectional GAN more suited to detect network intrusion attacks. Unlike existing GANs used in natural image processing which demands a strong dependence between the generator and the discriminator to produce realistic fake images, our proposed model allows the generator and the discriminator to be trained without needing to be in sync in their training iterations. By relaxing the dependence between these two, the generator and encoder, by working together, can produce more accurate synthetic text-based network traffic samples. In addition, by allowing more training of the generator and the encoder, our model produces the fake datasets with its underlying semantic representation intact similar to the real data.  

Our model is also equipped with a one-class binary classifier for the trained encoder and discriminator to use for detecting anomalous traffic from normal traffic. By offering a one-class (binary) classifier, a complex calculation involved in finding threshold or anomaly score can be avoided. 

Our proposed model, evaluated on extensive experimental results, demonstrates that it is highly effective for the generator to produce a synthetic network traffic dataset that can contribute to detect anomalous network traffic. Our benchmarking result shows that our proposed model outperformed other similar generative models by achieving more than 92\% F1-score. 

We plan to extend our work as a general data argumentation engine to produce a synthetic dataset for minority classes often used in DDoS attack classification \cite{wei2021ae}, Android-based malware detection \cite{zhu2020multi, zhu2021task}, or ransomware detection and classification tasks \cite{zhu2021few, mcintosh2018large, mcintosh2019inadequacy}. We also plan to apply our technique for other application area, such as finding detects in x-ray images \cite{feng2022automated} to evaluate the feasibility, extensionability, and generalizability of our approach.

\section*{Acknowledgment}
This work is supported by the Cyber Security Research Programme—Artificial Intelligence for Automating Response to Threats from the Ministry of Business, Innovation, and Employment (MBIE) of New Zealand as a part of the Catalyst Strategy Funds under the grant number MAUX1912.

\bibliographystyle{unsrtnat}
\bibliography{mybib} 

\begin{thebibliography}{31}
\providecommand{\natexlab}[1]{#1}
\providecommand{\url}[1]{\texttt{#1}}
\expandafter\ifx\csname urlstyle\endcsname\relax
  \providecommand{\doi}[1]{doi: #1}\else
  \providecommand{\doi}{doi: \begingroup \urlstyle{rm}\Url}\fi

\bibitem[Jang-Jaccard and Nepal(2014)]{jang2014survey}
Julian Jang-Jaccard and Surya Nepal.
\newblock A survey of emerging threats in cybersecurity.
\newblock \emph{Journal of Computer and System Sciences}, 80\penalty0
  (5):\penalty0 973--993, 2014.

\bibitem[Ahmad et~al.(2021)Ahmad, Shahid~Khan, Wai~Shiang, Abdullah, and
  Ahmad]{ett.4150}
Zeeshan Ahmad, Adnan Shahid~Khan, Cheah Wai~Shiang, Johari Abdullah, and Farhan
  Ahmad.
\newblock Network intrusion detection system: A systematic study of machine
  learning and deep learning approaches.
\newblock \emph{Transactions on Emerging Telecommunications Technologies},
  32\penalty0 (1):\penalty0 e4150, 2021.
\newblock \doi{https://doi.org/10.1002/ett.4150}.

\bibitem[Zhu et~al.(2021{\natexlab{a}})Zhu, Jang-Jaccard, Liu, and
  Zhou]{zhu2021joint}
Jinting Zhu, Julian Jang-Jaccard, Tong Liu, and Jukai Zhou.
\newblock Joint spectral clustering based on optimal graph and feature
  selection.
\newblock \emph{Neural Processing Letters}, 53\penalty0 (1):\penalty0 257--273,
  2021{\natexlab{a}}.

\bibitem[Kingma and Welling(2013)]{kingma2013auto}
Diederik~P Kingma and Max Welling.
\newblock Auto-encoding variational bayes.
\newblock \emph{arXiv preprint arXiv:1312.6114}, 2013.

\bibitem[Goodfellow et~al.(2014)Goodfellow, Pouget-Abadie, Mirza, Xu,
  Warde-Farley, Ozair, Courville, and Bengio]{goodfellow2014generative}
Ian Goodfellow, Jean Pouget-Abadie, Mehdi Mirza, Bing Xu, David Warde-Farley,
  Sherjil Ozair, Aaron Courville, and Yoshua Bengio.
\newblock Generative adversarial nets.
\newblock \emph{Advances in neural information processing systems}, 27, 2014.

\bibitem[Javaid et~al.(2016)Javaid, Niyaz, Sun, and Alam]{deep2}
Ahmad Javaid, Quamar Niyaz, Weiqing Sun, and Mansoor Alam.
\newblock A deep learning approach for network intrusion detection system.
\newblock \emph{Eai Endorsed Transactions on Security and Safety}, 3\penalty0
  (9):\penalty0 e2, 2016.

\bibitem[An and Cho(2015)]{an2015variational}
Jinwon An and Sungzoon Cho.
\newblock Variational autoencoder based anomaly detection using reconstruction
  probability.
\newblock \emph{Special Lecture on IE}, 2\penalty0 (1):\penalty0 1--18, 2015.

\bibitem[Chang et~al.(2020)Chang, Tu, Xie, and
  Yuan]{10.1007/978-3-030-58555-6_20}
Yunpeng Chang, Zhigang Tu, Wei Xie, and Junsong Yuan.
\newblock Clustering driven deep autoencoder for video anomaly detection.
\newblock In Andrea Vedaldi, Horst Bischof, Thomas Brox, and Jan-Michael Frahm,
  editors, \emph{Computer Vision -- ECCV 2020}, pages 329--345, Cham, 2020.
  Springer International Publishing.
\newblock ISBN 978-3-030-58555-6.

\bibitem[Xu et~al.(2021)Xu, Jang-Jaccard, Singh, Wei, and Sabrina]{9552882}
Wen Xu, Julian Jang-Jaccard, Amardeep Singh, Yuanyuan Wei, and Fariza Sabrina.
\newblock Improving performance of autoencoder-based network anomaly detection
  on nsl-kdd dataset.
\newblock \emph{IEEE Access}, 9:\penalty0 140136--140146, 2021.
\newblock \doi{10.1109/ACCESS.2021.3116612}.

\bibitem[Sadaf and Sultana(2020)]{9189883}
Kishwar Sadaf and Jabeen Sultana.
\newblock Intrusion detection based on autoencoder and isolation forest in fog
  computing.
\newblock \emph{IEEE Access}, 8:\penalty0 167059--167068, 2020.
\newblock \doi{10.1109/ACCESS.2020.3022855}.

\bibitem[Aygun and Yavuz(2017)]{7987197}
R.~Can Aygun and A.~Gokhan Yavuz.
\newblock Network anomaly detection with stochastically improved autoencoder
  based models.
\newblock In \emph{2017 IEEE 4th International Conference on Cyber Security and
  Cloud Computing (CSCloud)}, pages 193--198, 2017.
\newblock \doi{10.1109/CSCloud.2017.39}.

\bibitem[Doersch(2016)]{doersch2016tutorial}
Carl Doersch.
\newblock Tutorial on variational autoencoders.
\newblock \emph{arXiv preprint arXiv:1606.05908}, 2016.

\bibitem[Schlegl et~al.(2017)Schlegl, Seeb{\"o}ck, Waldstein, Schmidt-Erfurth,
  and Langs]{schlegl2017unsupervised}
Thomas Schlegl, Philipp Seeb{\"o}ck, Sebastian~M Waldstein, Ursula
  Schmidt-Erfurth, and Georg Langs.
\newblock Unsupervised anomaly detection with generative adversarial networks
  to guide marker discovery.
\newblock In \emph{International conference on information processing in
  medical imaging}, pages 146--157. Springer, 2017.

\bibitem[Schlegl et~al.(2019)Schlegl, Seeböck, Waldstein, Langs, and
  Schmidt-Erfurth]{SCHLEGL201930}
Thomas Schlegl, Philipp Seeböck, Sebastian~M. Waldstein, Georg Langs, and
  Ursula Schmidt-Erfurth.
\newblock f-anogan: Fast unsupervised anomaly detection with generative
  adversarial networks.
\newblock \emph{Medical Image Analysis}, 54:\penalty0 30--44, 2019.
\newblock ISSN 1361-8415.
\newblock \doi{https://doi.org/10.1016/j.media.2019.01.010}.

\bibitem[Akcay et~al.(2019)Akcay, Atapour-Abarghouei, and
  Breckon]{10.1007/978-3-030-20893-6_39}
Samet Akcay, Amir Atapour-Abarghouei, and Toby~P. Breckon.
\newblock Ganomaly: Semi-supervised anomaly detection via adversarial training.
\newblock In C.~V. Jawahar, Hongdong Li, Greg Mori, and Konrad Schindler,
  editors, \emph{Computer Vision -- ACCV 2018}, pages 622--637, Cham, 2019.
  Springer International Publishing.
\newblock ISBN 978-3-030-20893-6.

\bibitem[Zenati et~al.(2018{\natexlab{a}})Zenati, Foo, Lecouat, Manek, and
  Chandrasekhar]{zenati2018efficient}
Houssam Zenati, Chuan~Sheng Foo, Bruno Lecouat, Gaurav Manek, and
  Vijay~Ramaseshan Chandrasekhar.
\newblock Efficient gan-based anomaly detection.
\newblock \emph{arXiv preprint arXiv:1802.06222}, 2018{\natexlab{a}}.

\bibitem[Kaplan and Alptekin(2020)]{KAPLAN2020185}
M.~Oguz Kaplan and S.~Emre Alptekin.
\newblock An improved bigan based approach for anomaly detection.
\newblock \emph{Procedia Computer Science}, 176:\penalty0 185--194, 2020.
\newblock ISSN 1877-0509.
\newblock \doi{https://doi.org/10.1016/j.procs.2020.08.020}.

\bibitem[Mohammadi and Sabokrou(2019)]{8990759}
Bahram Mohammadi and Mohammad Sabokrou.
\newblock End-to-end adversarial learning for intrusion detection in computer
  networks.
\newblock In \emph{2019 IEEE 44th Conference on Local Computer Networks (LCN)},
  pages 270--273, 2019.
\newblock \doi{10.1109/LCN44214.2019.8990759}.

\bibitem[Dumoulin et~al.(2016)Dumoulin, Belghazi, Poole, Mastropietro, Lamb,
  Arjovsky, and Courville]{dumoulin2016adversarially}
Vincent Dumoulin, Ishmael Belghazi, Ben Poole, Olivier Mastropietro, Alex Lamb,
  Martin Arjovsky, and Aaron Courville.
\newblock Adversarially learned inference.
\newblock \emph{arXiv preprint arXiv:1606.00704}, 2016.

\bibitem[Donahue et~al.(2016)Donahue, Kr{\"a}henb{\"u}hl, and
  Darrell]{donahue2017adversarial}
Jeff Donahue, Philipp Kr{\"a}henb{\"u}hl, and Trevor Darrell.
\newblock Adversarial feature learning.
\newblock \emph{arXiv preprint arXiv:1605.09782}, 2016.

\bibitem[Arjovsky and Bottou(2017)]{arjovsky2017principled}
Martin Arjovsky and Léon Bottou.
\newblock Towards principled methods for training generative adversarial
  networks, 2017.

\bibitem[Berthelot et~al.(2017)Berthelot, Schumm, and Metz]{berthelot2017began}
David Berthelot, Thomas Schumm, and Luke Metz.
\newblock Began: Boundary equilibrium generative adversarial networks.
\newblock \emph{arXiv preprint arXiv:1703.10717}, 2017.

\bibitem[Ieracitano et~al.(2020)Ieracitano, Adeel, Morabito, and
  Hussain]{IERACITANO202051}
Cosimo Ieracitano, Ahsan Adeel, Francesco~Carlo Morabito, and Amir Hussain.
\newblock A novel statistical analysis and autoencoder driven intelligent
  intrusion detection approach.
\newblock \emph{Neurocomputing}, 387:\penalty0 51--62, 2020.
\newblock ISSN 0925-2312.
\newblock \doi{https://doi.org/10.1016/j.neucom.2019.11.016}.

\bibitem[Zenati et~al.(2018{\natexlab{b}})Zenati, Romain, Foo, Lecouat, and
  Chandrasekhar]{8594897}
Houssam Zenati, Manon Romain, Chuan-Sheng Foo, Bruno Lecouat, and Vijay
  Chandrasekhar.
\newblock Adversarially learned anomaly detection.
\newblock In \emph{2018 IEEE International Conference on Data Mining (ICDM)},
  pages 727--736, 2018{\natexlab{b}}.
\newblock \doi{10.1109/ICDM.2018.00088}.

\bibitem[Chen and Jiang(2019)]{chen2019efficient}
Hongyu Chen and Li~Jiang.
\newblock Efficient gan-based method for cyber-intrusion detection.
\newblock \emph{arXiv preprint arXiv:1904.02426}, 2019.

\bibitem[Wei et~al.(2021)Wei, Jang-Jaccard, Sabrina, Singh, Xu, and
  Camtepe]{wei2021ae}
Yuanyuan Wei, Julian Jang-Jaccard, Fariza Sabrina, Amardeep Singh, Wen Xu, and
  Seyit Camtepe.
\newblock Ae-mlp: A hybrid deep learning approach for ddos detection and
  classification.
\newblock \emph{IEEE Access}, 9:\penalty0 146810--146821, 2021.

\bibitem[Zhu et~al.(2020)Zhu, Jang-Jaccard, and Watters]{zhu2020multi}
Jinting Zhu, Julian Jang-Jaccard, and Paul~A Watters.
\newblock Multi-loss siamese neural network with batch normalization layer for
  malware detection.
\newblock \emph{IEEE Access}, 8:\penalty0 171542--171550, 2020.

\bibitem[Zhu et~al.(2021{\natexlab{b}})Zhu, Jang-Jaccard, Singh, Watters, and
  Camtepe]{zhu2021task}
Jinting Zhu, Julian Jang-Jaccard, Amardeep Singh, Paul~A Watters, and Seyit
  Camtepe.
\newblock Task-aware meta learning-based siamese neural network for classifying
  obfuscated malware.
\newblock \emph{arXiv preprint arXiv:2110.13409}, 2021{\natexlab{b}}.

\bibitem[Zhu et~al.(2021{\natexlab{c}})Zhu, Jang-Jaccard, Singh, Welch,
  AI-Sahaf, and Camtepe]{zhu2021few}
Jinting Zhu, Julian Jang-Jaccard, Amardeep Singh, Ian Welch, Harith AI-Sahaf,
  and Seyit Camtepe.
\newblock A few-shot meta-learning based siamese neural network using entropy
  features for ransomware classification.
\newblock \emph{arXiv preprint arXiv:2112.00668}, 2021{\natexlab{c}}.

\bibitem[McIntosh et~al.(2018)McIntosh, Jang-Jaccard, and
  Watters]{mcintosh2018large}
Timothy~R McIntosh, Julian Jang-Jaccard, and Paul~A Watters.
\newblock Large scale behavioral analysis of ransomware attacks.
\newblock In \emph{International Conference on Neural Information Processing},
  pages 217--229. Springer, 2018.

\bibitem[McIntosh et~al.(2019)McIntosh, Jang-Jaccard, Watters, and
  Susnjak]{mcintosh2019inadequacy}
Timothy McIntosh, Julian Jang-Jaccard, Paul Watters, and Teo Susnjak.
\newblock The inadequacy of entropy-based ransomware detection.
\newblock In \emph{International Conference on Neural Information Processing},
  pages 181--189. Springer, 2019.

\end{thebibliography}






\end{document}